\documentclass{article}

\usepackage[position,final]{neurips_2026}
\usepackage{microtype}
\usepackage{graphicx}
\usepackage{subcaption}
\usepackage{booktabs} 
\usepackage[utf8]{inputenc}
\usepackage{booktabs}   
\usepackage{multirow}   
\usepackage{amssymb}    
\usepackage{adjustbox}  
\usepackage{xcolor}     
\usepackage{pifont}
\usepackage[table]{xcolor}
\usepackage{hyperref}
\usepackage{tabularx}   
\usepackage{makecell}
\usepackage[utf8]{inputenc} 
\usepackage[T1]{fontenc}    
\usepackage{hyperref}       
\usepackage{url}            
\usepackage{booktabs}       
\usepackage{amsfonts}       
\usepackage{nicefrac}       
\usepackage{microtype}      
\usepackage{xcolor}         

\usepackage{amsmath}
\usepackage{amssymb}
\usepackage{mathtools}
\usepackage{amsthm}

\usepackage[capitalize,noabbrev]{cleveref}

\theoremstyle{plain}

\theoremstyle{definition}

\theoremstyle{remark}

\usepackage[textsize=tiny]{todonotes}
\definecolor{mygreen}{rgb}{0, 0.6, 0}
\definecolor{myred}{rgb}{0.8, 0, 0}
\definecolor{metricgray}{gray}{0.92}

\newcommand{\ok}{{\color{mygreen}\ding{51}}} 
\newcommand{\no}{{\color{myred}\ding{55}}}  
\newcommand{\gok}{\cellcolor{metricgray}\ok}
\newcommand{\gno}{\cellcolor{metricgray}\no}

\title{Evaluation of Visual Processing Should Be Human-Centered, Not Metric-Centered}

\author{%
  Fanghua Yu$^{1,2,*}$,
  Jinfan Hu$^{3,4,*}$,
  Zhiyuan You$^{5}$,
  Xiang Yin$^{3}$,
  Hongyu An$^{6}$, \\
    \bfseries
  Xinqi Lin$^{3,4}$,
  Hongyang Li$^{1}$,
  Chao Dong$^{3,7,\dagger}$,
  Jinjin Gu$^{6,\dagger}$ \\[0.5em]
  \normalfont
  \footnotesize
  $^{1}$~The University of Hong Kong \quad
  \footnotesize
  $^{2}$~Shenzhen Loop Area Institute \\
  \footnotesize
  $^{3}$~Shenzhen Institutes of Advanced Technology, Chinese Academy of Sciences \\
  \footnotesize
  $^{4}$~University of Chinese Academy of Sciences \quad
  \footnotesize
  $^{5}$~The Chinese University of Hong Kong \\
  \footnotesize
  $^{6}$~INSAIT, Sofia University ``St.~Kliment Ohridski'' \quad
  \footnotesize
  $^{7}$~Shenzhen University of Advanced Technology \quad
 \\[0.3em]
  \footnotesize
  $^{*}$Equal contribution \quad
  $^{\dagger}$Corresponding authors \quad
  \texttt{\{chao.dong@siat.ac.cn, jinjin.gu@insait.ai\}}
}

\begin{document}

\maketitle
\begin{abstract}

This position paper argues that the evaluation of modern visual processing systems should no longer be driven primarily by single-metric image quality assessment benchmarks, particularly in the era of generative and perception-oriented methods. Image restoration exemplifies this divergence: while objective IQA metrics enable reproducible, scalable evaluation, they have increasingly drifted apart from human perception and user preferences. We contend that this mismatch risks constraining innovation and misguiding research progress across visual processing tasks. Rather than rejecting metrics altogether, this paper calls for a rebalancing of evaluation paradigms, advocating a more human-centered, context-aware, and fine-grained approach to assessing the visual models' outcomes.
\end{abstract}

\section{Introduction}

\label{sec:intro}
Since the introduction of deep learning into computer vision, intelligent models for visual processing tasks, particularly in image restoration and super resolution, have advanced rapidly.
These approaches have gradually established deep learning as the dominant paradigm in the field, continuously driving performance improvements across a wide range of standard benchmark datasets.
For a long period of time, and even to the present day, performance on image quality assessment (IQA) benchmarks has remained one of the primary motivations for proposing new algorithms, as well as the most central criterion for paper evaluation, peer review, and method comparison.
Under this research paradigm, progress in the field is often externalized as incremental improvements on benchmark metrics, where PSNR, SSIM, and LPIPS remain the main optimization targets for image restoration models.

However, it has become increasingly evident that the guiding power of this objective evaluation framework has significantly weakened.
In particular, with the rise of perceptually driven image super-resolution and image processing methods, and the introduction of generative models such as GANs and diffusion models, the objectives of image restoration have long extended beyond the minimization of pixel-level distortion.
While generative models substantially enhance perceptual quality through improved high-frequency detail synthesis, they also systematically weaken the correspondence between IQA metrics and human subjective perception.
A growing body of work \cite{tradeoff,pipal,NTIRE2022,SUPIR} demonstrates that algorithms achieving superior numerical performance on mainstream benchmarks do not necessarily produce results that are more visually preferable in real-world usage, and may even exhibit perceptual degradation.

At the same time, image processing methods driven by generative models have begun to demonstrate a higher level of intelligence.
Rather than producing a single deterministic output, these methods are capable of generating diverse results with different stylistic biases and visual preferences while preserving semantic consistency.
Such behavior better reflects the diversity of real user preferences and aesthetic expectations.
Nevertheless, evaluation paradigms centered around objective IQA scores impose implicit constraints on this diversity and personalization.
Overemphasis on uniform benchmark performance inevitably suppresses algorithmic explorations that may better satisfy heterogeneous user needs but do not perform optimally under metrics.

In other words, in the era of generative image restoration, relying solely on IQA metrics to measure progress is no longer sufficient and lacks the necessary granularity.
The deep mismatch between academic metrics and real audience experience has become an unavoidable issue in many subfields.
Despite its far-reaching implications, this problem has yet to receive systematic and sustained attention from the community.
This position paper is devoted to discussing this mismatch and calls for a reexamination of evaluation paradigms in image restoration.
\textbf{We argue that the current evaluation paradigm has become a fundamental bottleneck, as it misaligns design objectives with human preferences and impedes the development of more versatile, human-centric image restoration.}
\vspace{-3mm}

\begin{figure}[t]
    \centering
    
    \begin{minipage}[t]{0.484\linewidth}
        \centering
        \includegraphics[width=\linewidth]{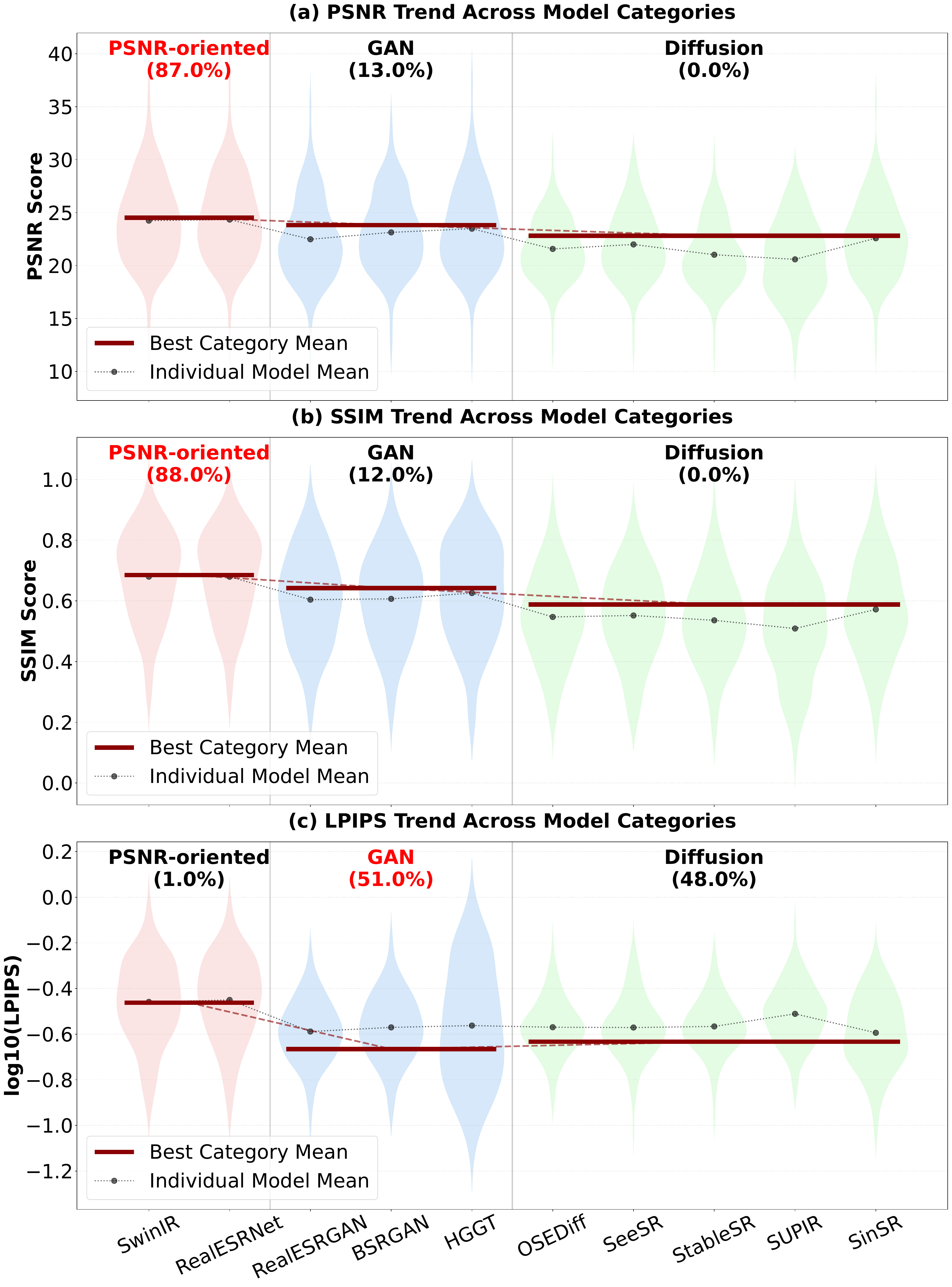}
        \vspace{-5mm}
        \captionof{figure}{Trend of full-reference IQA metrics (PSNR, SSIM, LPIPS). The percentages shown below each category represent the corresponding winning rates among those categories. ``Best Category Mean" represents the mean value of the optimal model performance of each category.}
        \label{fig:fr-iqa trend}
    \end{minipage}
    \hfill
    \begin{minipage}[t]{0.49\linewidth}
        \centering
        \includegraphics[width=\linewidth]{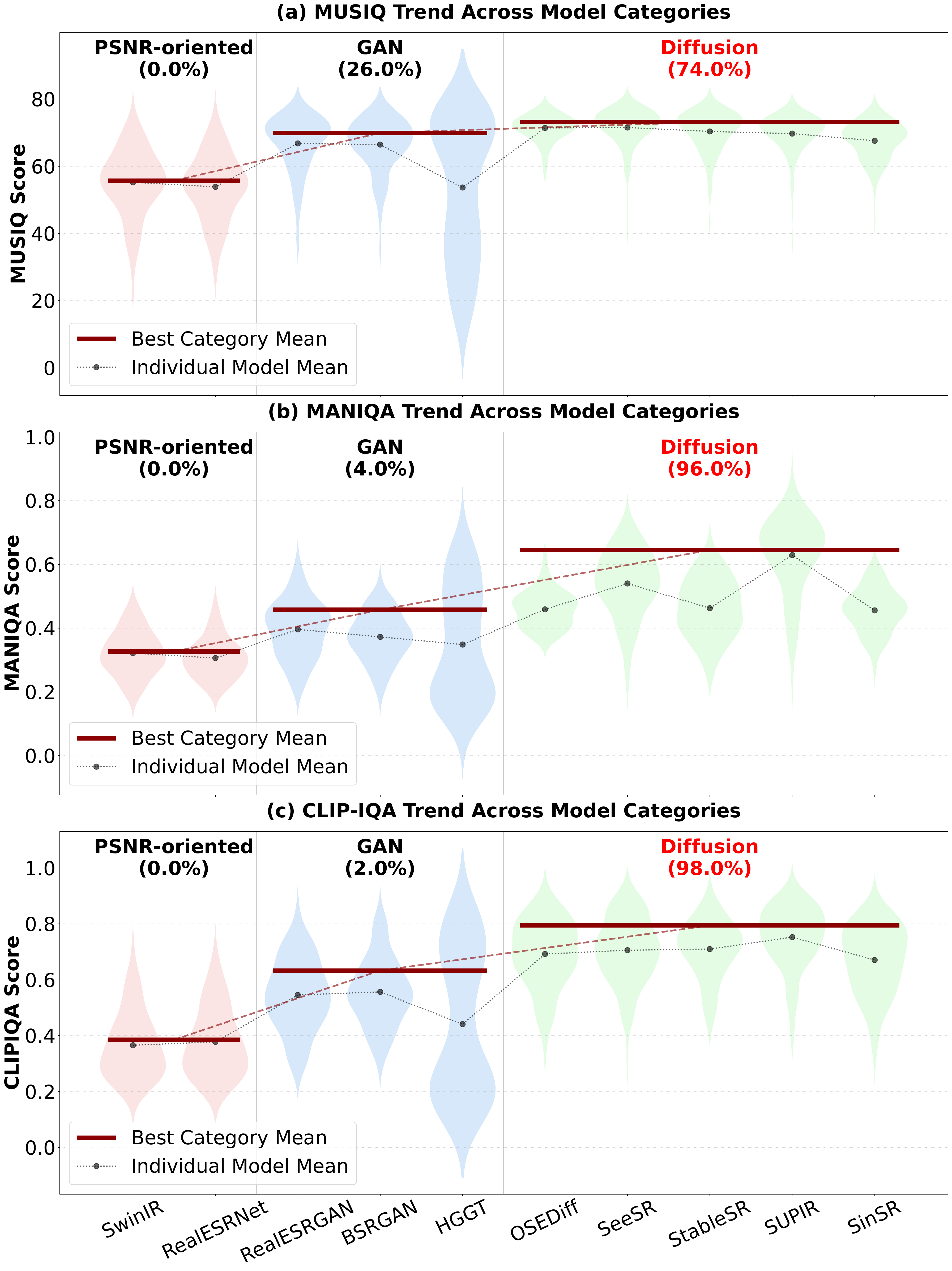}
        \vspace{-5mm}
        \captionof{figure}{Trend of NR-IQA metrics (MUSIQ, MANIQA, CLIP-IQA). The percentages shown below each category represent the corresponding winning rates among those categories. ``Best Category Mean" represents the mean value of the optimal model performance of each category.}
        \label{fig:nr-iqa trend}
    \end{minipage}
    \vspace{-7mm}
\end{figure}

\subsection{Our position}

Based on the observations above, this paper takes the position that image processing research should no longer be guided by evaluation metrics as the sole objective.
We must acknowledge a practical reality of the current academic ecosystem: most studies are still pressured to emphasize numerical superiority on existing benchmarks in order to be publishable.
This situation does not stem from conservatism on the part of individual researchers, but rather from long-standing collective assumptions and evaluation inertia within the field.
The core issue is that the design principles and expressive capacity of prevailing IQA metrics have lagged far behind the rapid evolution of generative image processing methods.
Because peer review and publication decisions continue to rely heavily on these metrics, researchers are forced to invest substantial effort into metric-oriented optimization, often at the cost of exploring more innovative directions.
Accordingly, our first central position is a call for a more open and cautious attitude toward the role of evaluation metrics in image processing research.
Metrics should not be treated as absolute arbiters of progress, but instead interpreted in the context of specific tasks, methodological properties, and application scenarios.
This is particularly critical for generative image restoration tasks, which are inherently nondeterministic, where simple numerical comparisons frequently fail to capture meaningful differences in model behavior.

Second, we argue that evaluation objectives should be explicitly differentiated across application domains.
For safety-critical and rigorously constrained fields such as medical imaging, remote sensing, and radar signal processing, evaluation criteria and optimization goals must be considered separately and with greater caution.
In contrast, for image processing tasks primarily serving human visual experience and entertainment consumption, the fundamental evaluation criterion is inherently human-centered.
Perceptual quality, aesthetic preference, and stylistic coherence are ultimately judged by human users rather than by metrics.

As image processing research continues to enter deeper and more mature stages, it becomes increasingly important to move beyond isolated model performance comparisons and to consider the broader real-world context in which these methods operate.
Factors such as deployability, human interaction, and user preference diversity often play a decisive role in practical success, yet they remain largely overlooked by existing evaluation paradigms.

Under this premise, we contend that both evaluation frameworks and method design should adopt human perception and preference as their guiding principle.
A key challenge for the community is how to systematically incorporate human evaluation in a way that is scalable, reproducible, analyzable, and comparable across studies, rather than treating it as a one-off supplementary experiment.
Addressing this challenge should be regarded as a core research topic in contemporary image processing.

Finally, we do not deny the importance of metric-based research in the advancement of the field.
Objective evaluation metrics remain indispensable tools for model selection, large-scale comparison, and rapid iteration, and many practical systems must still rely on such metrics for training, fine-tuning, and assessment.
Precisely for this reason, the development of IQA itself deserves substantially greater research attention. 
Future efforts should focus on enhancing the perceptual alignment of metrics with human, and even on extending their capabilities beyond those of current image restoration techniques.

On the one hand, existing IQA methods must better accommodate and reflect modern generative image processing paradigms.
Many current IQA datasets lag behind state-of-the-art generative models in terms of resolution, content complexity, and diversity of synthesized outputs.
On the other hand, IQA should evolve toward more fine-grained and multi-dimensional evaluation forms.
For example, by combining Q series \cite{Q-Align,Q-Bench,Q-instruct,Q-Bench2,Q-ground,Q-Insight} and DepictQA \cite{DepictQA, DepictQA-wild} style approaches with large language models and vision language models capable of semantic-level image understanding, it becomes possible to produce more interpretable and behavior-aware assessments of generative image processing methods.
This situation is further exacerbated by the long-standing separation between the IQA research community and the broader image processing community, which renders cross-community feedback indirect, costly, and significantly delayed.

We argue that future image quality evaluation should not be confined to single-score comparisons, but should instead provide structured, multidimensional, and explainable feedback signals.
This constitutes one of the central positions that this paper aims to articulate to the community.

\subsection{Alternative Views}

A natural counterargument to the position advocated in this paper is that metric-centered evaluation remains essential for scientific rigor, scalability, and cumulative progress in image processing research. Objective metrics enable reproducible comparison across methods, datasets, and time, and serve as a common language that allows large-scale collaboration within the whole community. 

We acknowledge the validity of this view. Objective metrics have played a crucial role in the historical development of digital image processing and remain indispensable for many practical workflows, including large-scale training, ablation analysis, and rapid iteration. Moreover, human evaluation is often expensive, time-consuming, and difficult to standardize, which limits its applicability as a primary evaluation tool in many settings.

A related concern is that human preferences are inherently noisy and inconsistent. Subjective judgments vary across individuals, cultures, and contexts, and may fluctuate for users over time. From this standpoint, centrally elevating human evaluation risks introducing unstable and irreproducible signals into assessment, undermining the reliability objective metrics are designed to provide.

Finally, it is important to note that not all image processing tasks are perception-driven. In safety-critical domains such as medical imaging, remote sensing, and scientific visualization, fidelity to physical measurements and numerical accuracy often take precedence over perceptual appeal.

While we agree with these views in their respective contexts, we argue that they do not negate the core concern raised in this paper. Rather than rejecting metric-based evaluation, our position is that its dominance has expanded beyond its scope, particularly in perception-oriented and generative image processing tasks. The challenge, therefore, is not to replace objective metrics but to reconsider their role, limitations, and interaction with human-centered evaluation for future directions.

\section{The Risks of Metric-Centric Research}

\label{sec:metric_centric}

\subsection{IQA Metrics as De Facto Research Objectives}

It is common in image restoration to report multiple IQA metrics in comparison tables to highlight a method's advantages. %
Such reporting enables standardized benchmarking and a shared quantitative reference for the community.
However, it can also unintentionally shift attention toward numerical improvements as a primary indicator of research success.
As a result, benchmark scores may gradually evolve from diagnostic tools into de facto research objectives.

This metric-centric culture extends beyond benchmarking into publication and peer-review processes.
Numerical improvements often become central decision criteria when assessing technical contributions.
Researchers may therefore feel pressure to allocate substantial effort toward optimizing metric performance rather than improving perceptual quality or user experience.
Such incentive structures can bias research directions toward what is easily measurable instead of what is genuinely beneficial.

More importantly, improvements (or declines) in these quantitative scores do not always fully reflect the actual technological progress of restoration models.
As restoration systems become more capable of synthesizing realistic and semantically meaningful content, their behavior increasingly departs from assumptions embedded in traditional Full-Reference (FR) image quality assessment.
This divergence makes it difficult for existing metrics to represent perceptual quality or realism faithfully.

Together, these issues suggest that leaderboard-driven evaluation may oversimplify research progress and system capability, potentially steering the field in misleading directions. 
This discrepancy motivates a closer examination of the assumptions and limitations of current IQA methodologies.
%


\subsection{The Gap Between Metrics and Perceptual Quality}

\textbf{Before GANs (Deterministic Restoration).}
Early restoration methods \cite{SRCNN,CSC-SR,VDSR} aimed to approximate a single distortion-free solution. Under this setting, pixel-wise metrics were well aligned with the modeling objective, since the task was largely treated as regression toward a reference image.
\textbf{GAN Era (Realistic Detail Synthesis).}
With the introduction of GAN-based restoration \cite{SRGAN}, models \cite{ESRGAN,Ranksrgan,Real-ESRGAN} began generating perceptually plausible details that no longer strictly correspond to the reference signal. This shift weakened the assumptions behind distance-based metrics. Benchmarks such as PIPAL \cite{pipal} exposed the increasing divergence between human preference and similarity-driven evaluation.
\textbf{Diffusion Era (Semantic Reconstruction).}
Diffusion-based restoration \cite{S3,Diffir,DiffBIR,stablesr,SeeSR} further expands this trend by enabling semantically consistent and realistic content generation. Restoration is no longer limited to recovering lost signals but often involves reconstructing plausible structures and textures. Nevertheless, many IQA methodologies still operate under assumptions closer to earlier regimes, leaving a widening gap between evaluation protocols and model capabilities.

The evolution of image restoration models has fundamentally challenged the traditional FR IQA methodologies.
The SRIQA benchmark \cite{sriqa} comprises restoration outputs from diverse models across 100 images.
We conduct comprehensive IQA metric calculations for these outputs, with the results presented in \figurename~\ref{fig:fr-iqa trend}.
Traditional metrics such as PSNR and SSIM \cite{ssim} have become increasingly inadequate for evaluating modern image restoration systems.
As restoration methods evolve, improvements in perceptual realism are often accompanied by declining PSNR/SSIM scores, not because models are worse, but because these metrics favor pixel-wise similarity and penalize semantically plausible deviations from the reference.
LPIPS \cite{lpips} partially addressed this issue in the GAN era by measuring deep feature similarity rather than raw pixel error.
It could better distinguish PSNR-oriented methods from GAN-based approaches.
PSNR-oriented methods tend to produce over-smoothed textures due to regression toward a pixel-space average. 
In contrast, GAN-based methods relax strict pixel-level constraints and are able to synthesize high-frequency details.
However, LPIPS still primarily responds to high-frequency variations and does not understand whether synthesized textures are semantically or physically plausible.

\begin{figure}[t]
    \centering
    
    \begin{minipage}[t]{0.51\linewidth}
        \centering
        \includegraphics[width=\linewidth]{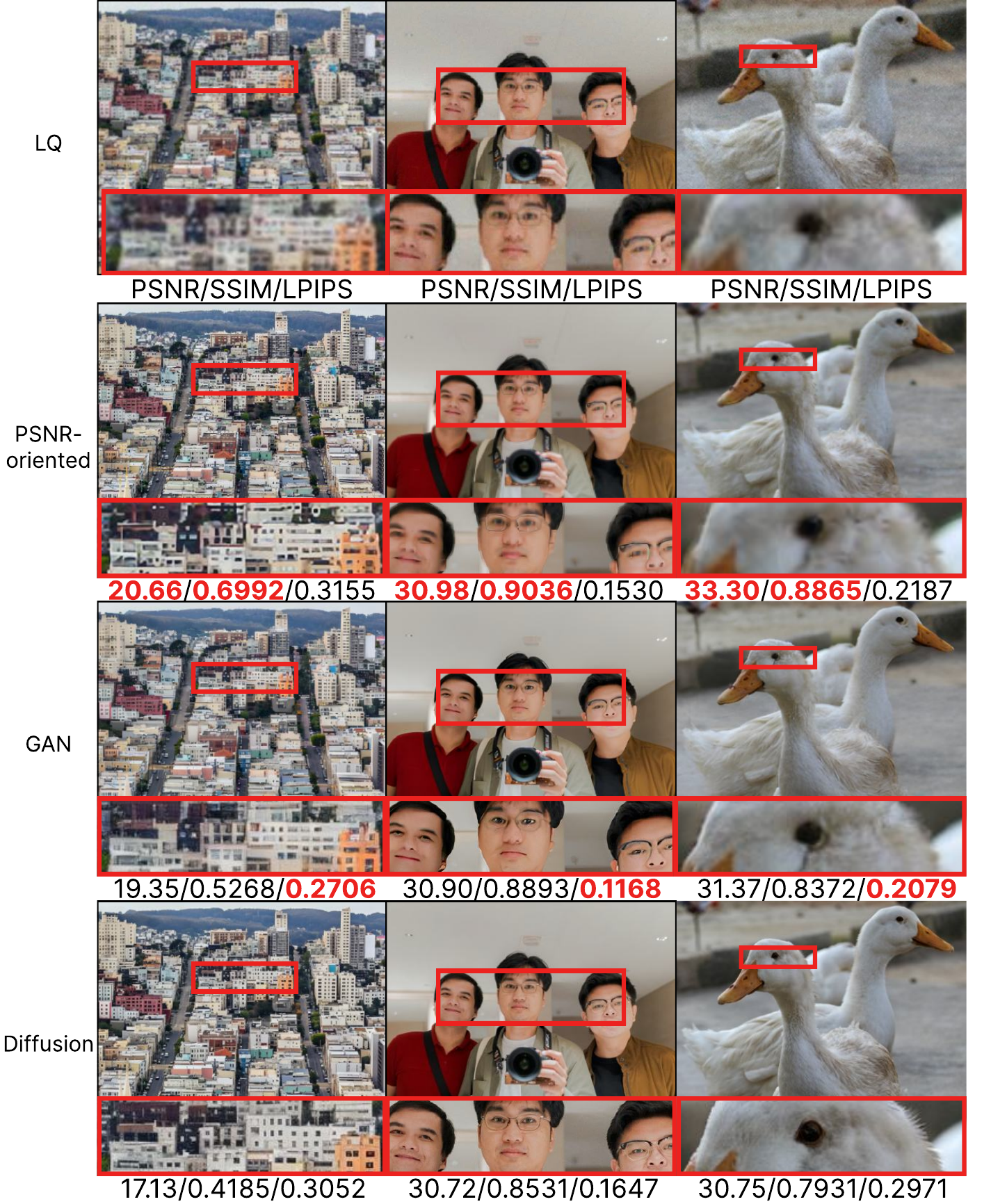}
        \vspace{-5mm}
        \captionof{figure}{PSNR, SSIM, and LPIPS metrics often misalign with perceptual quality. Red indicates the best result for each metric. Zoom in for details.}
        \label{fig:LPIPS_GAN}
    \end{minipage}
    \hfill
    \begin{minipage}[t]{0.47\linewidth}
        \centering
        \includegraphics[width=\linewidth]{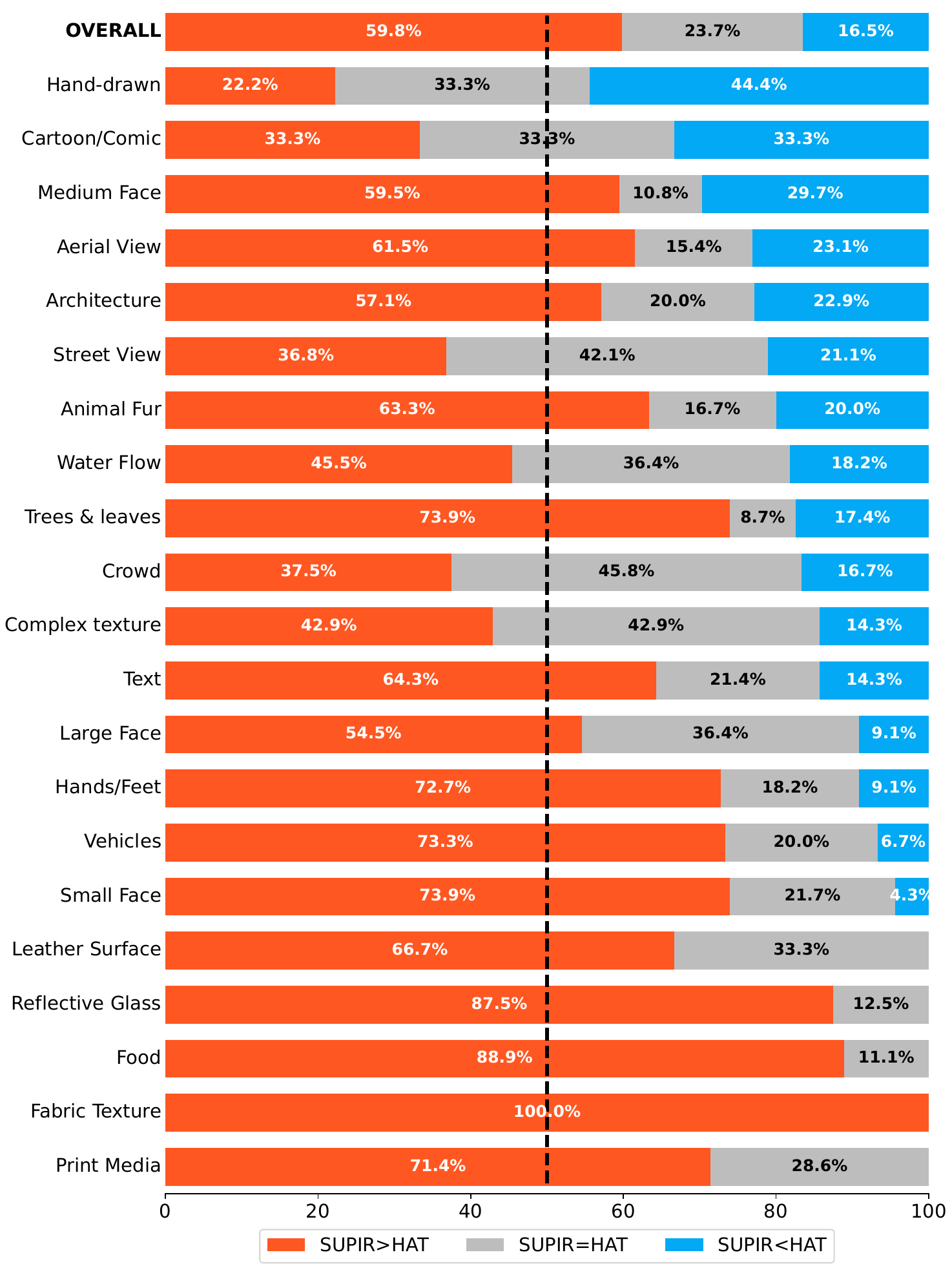}
        \vspace{-5mm}
        \captionof{figure}{User preference for model performance varies significantly across different semantic scenarios.}
        \label{fig:supir_vs_hat}
    \end{minipage}
    
    \vspace{-8mm}
\end{figure}



This limitation becomes more pronounced in the diffusion era. 
Diffusion-based restoration models generate richer and more coherent details, yet LPIPS may rate their outputs worse than GAN-based results, contradicting human perceptual judgments.
\figurename~\ref{fig:LPIPS_GAN} presents the visual performance of PSNR-oriented, GAN-based, and diffusion-based models across three typical scenarios (architecture, human faces, and animal fur). 
Quantitatively, PSNR-oriented models attain the highest values in both PSNR and SSIM metrics; yet visually, their outputs exhibit severe over-smoothing with a complete absence of fine details across all three scenarios. 
From the LPIPS evaluation perspective, GAN-based methods achieve better results in comparison with PSNR-oriented models.
Their outputs indeed contain richer fine details, avoid excessive smoothness, and render more well-defined facial structures, especially in the eye regions.
Nevertheless, the advantages of diffusion-based methods are immediately discernible upon visual inspection: they produce more distinct building contour differentiation, human faces that better conform to real-world anatomical structures, and exquisitely realistic fur textures on animals. 
However, none of the metrics, including PSNR, SSIM, and LPIPS, is capable of expressing the superior performance of diffusion-based methods.
Notably, LPIPS, explicitly designed for evaluating perceptual visual quality, nevertheless ranks GAN-based methods above diffusion-based ones.
This indicates a clear mismatch between numerical scores and actual perceptual quality.
It exposes a fundamental limitation of LPIPS: while it measures feature-level differences, it does not reliably reflect visual realism or semantic correctness.

\subsection{Exploitable Metrics and Inflated Scores}

In response to the growing mismatch between traditional metrics and modern restoration behavior, the community has increasingly turned to developing new perceptual quality assessment methods.
Recent efforts focus on learning-based No-Reference (NR) IQA models, which aim to evaluate visual quality without relying on a distortion-free reference image.
These approaches are designed to better align with human perception in scenarios where restoration outputs deviate from the original signal but appear visually plausible.

Representative examples include MUSIQ \cite{musiq}, MANIQA \cite{maniqa}, and CLIP-IQA \cite{clipiqa}.
Unlike earlier fidelity-oriented or shallow perceptual metrics, these methods attempt to capture richer structural, aesthetic, and semantic cues that are increasingly important in modern image restoration.

Using the SRIQA benchmark as a statistical testbed, we evaluated restoration methods across these three NR-IQA metrics and observed a consistent trend, as illustrated in \figurename~\ref{fig:nr-iqa trend}.
From PSNR-oriented models to GAN-based approaches and further to diffusion-based methods, the predicted perceptual quality scores show a steady increase along the technological progression.
This indicates that newer IQA models are more sensitive to the improvements introduced by recent restoration paradigms.

The performance improvement observed across model categories, progressing from PSNR-oriented methods to GAN-based and diffusion-based approaches, appears to be more consistent with human perceptual judgments.
This alignment may give the impression that recent learning-based IQA models are increasingly capable of evaluating the outputs of more advanced restoration systems.
In other words, these metrics seem to better capture the perceptual and semantic gains introduced by modern generative paradigms.

\begin{figure}[t]
    \centering
    
    \begin{minipage}[t]{0.49\linewidth}
        \centering
        \includegraphics[width=\linewidth]{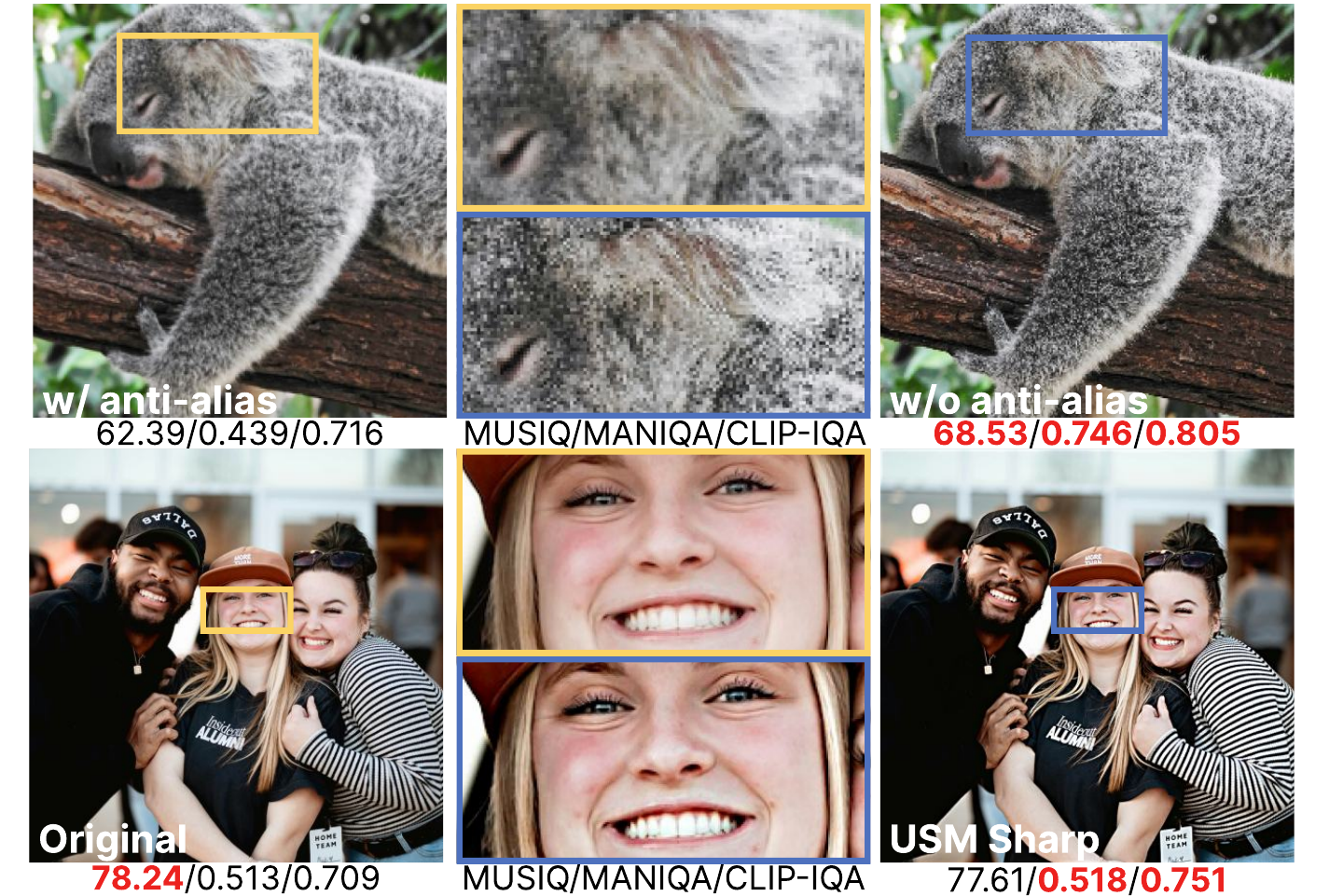}
        \vspace{-4mm}
        \captionof{figure}{Simple image manipulations can artificially boost NR IQA metrics, highlighting their vulnerability to manipulation.}
        \label{fig:hack_iqa}
    \end{minipage}
    \hfill
    \begin{minipage}[t]{0.49\linewidth}
        \centering
        \includegraphics[width=\linewidth]{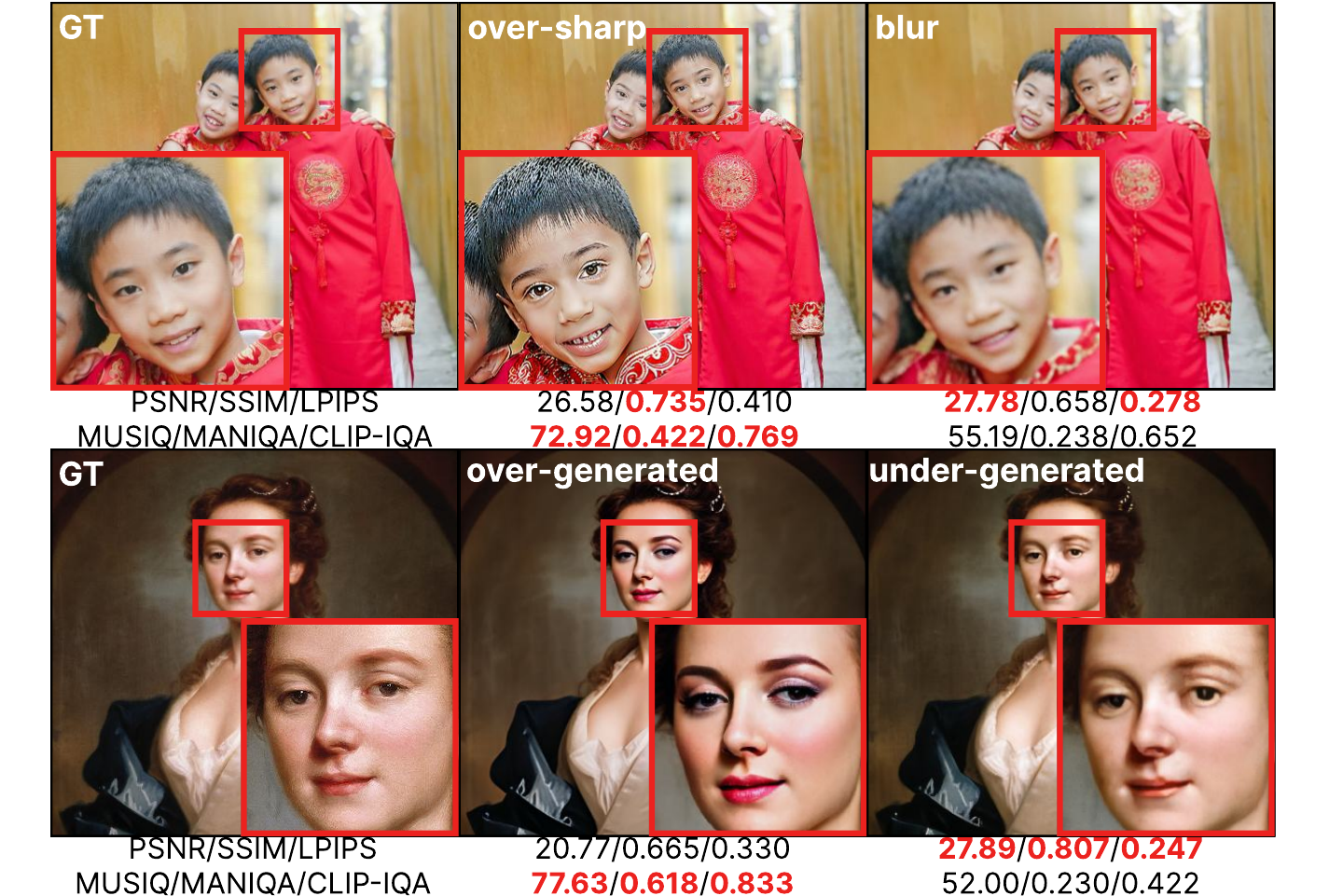}
        \vspace{-4mm}
        \captionof{figure}{Overall IQA scores fail to provide fine-grained evaluation; over-sharpening and over-generation can artificially inflate scores.}
        \label{fig:multi-dimension}
    \end{minipage}
    
    \vspace{-6pt}
\end{figure}

However, this encouraging trend also brings forward another critical issue that deserves careful attention: the overfitting risk of those learning-based IQA models.
A more direct observation is that many learning-based IQA metrics strongly favor images containing abundant high-frequency content.
Even when such content arises from over-sharpening, artificial enhancement, or irregular and cluttered textures, the corresponding IQA scores can still increase.
As illustrated in \figurename~\ref{fig:hack_iqa}, even simple processing tricks can artificially inflate IQA scores.
For example, choices as subtle as whether anti-aliasing is applied during downsampling, or applying Unsharp Masking (USM) to enhance high-frequency components, can lead to noticeable metric gains.
Furthermore, \figurename~\ref{fig:multi-dimension} shows that over-sharpened or over-generated results often achieve higher quantitative scores, despite appearing visually unpleasant or perceptually unnatural to human observers.
These phenomena indicate that certain metrics are sensitive to superficial high-frequency variations, which can be exploited without genuinely improving visual quality.
This suggests that these metrics may partially equate “more high-frequency variation” with “higher perceptual quality”, revealing a bias that can be exploited and does not always align with human visual judgment. 
In fact, this bias can incentivize researchers to introduce visually unnatural changes solely to inflate metric scores, rather than to produce results that are perceptually meaningful to humans.

\begin{table*}[tbp]
\centering
\renewcommand{\arraystretch}{0.05} 
\caption{Evolution and Adoption Analysis of IQA Metrics (2014--2025). Shaded cells (\colorbox{metricgray}{~}) indicate that the metric was available at the time of publication (IQA year $<$ Model year). The last row represents the IQA's adoption rate. *Adoption Rate = (Models reporting the metric) / (Models published after the metric's proposal).}
\label{tab:iqa-adoption-final}
\begin{adjustbox}{width=\textwidth,center}
\begin{tabular}{lcccccccccc} 
\toprule
\multirow{3}{*}{\textbf{Model}} & \multicolumn{3}{c}{\textbf{Full-Reference (FR)}} & \multicolumn{1}{c}{\textbf{Distribution}} & \multicolumn{4}{c}{\textbf{No-Reference (NR)}} & \multirow{3}{*}{\begin{tabular}[c]{@{}c@{}}\textbf{User}\\ \textbf{Study}\end{tabular}} & \multirow{3}{*}{\begin{tabular}[c]{@{}c@{}}\textbf{GitHub}\\ \textbf{Star}\end{tabular}}\\
\cmidrule(lr){2-4} \cmidrule(lr){5-5} \cmidrule(lr){6-9}
& \textbf{PSNR} & \textbf{SSIM} & \textbf{LPIPS} & \textbf{FID} & \textbf{NIQE} & \textbf{MUSIQ} & \textbf{MANIQA} & \textbf{CLIP-IQA} \\ 
& \textit{(Classic)} & \cite{ssim} & \cite{lpips} & \cite{fid} & \cite{niqe} & \cite{musiq} & \cite{maniqa} & \cite{clipiqa} & \\
\midrule
SRCNN \cite{SRCNN}  & \gok & \gno & $/$  & $/$  & \gno & $/$  & $/$  & $/$  & \gno & $/$\\
CSC-SR \cite{CSC-SR}   & \gok & \gno & $/$  & $/$  & \gno & $/$  & $/$  & $/$  & \gno & $/$\\
VDSR \cite{VDSR}  & \gok & \gok & $/$  & $/$  & \gno & $/$  & $/$  & $/$  & \gno & $/$\\
SRGAN \cite{SRGAN}  & \gok & \gok & $/$  & $/$  & \gno & $/$  & $/$  & $/$  & \gok & $/$\\
EDSR \cite{EDSR}  & \gok & \gok & $/$  & $/$  & \gno & $/$  & $/$  & $/$  & \gno & 2.6k\\
ESRGAN \cite{ESRGAN}  & \gok & \gno & $/$  & \gno & \gno & $/$  & $/$  & $/$  & \gno & 6.5k\\ 
MSRN \cite{MSRN}   & \gok & \gok & $/$  & \gno & \gno & $/$  & $/$  & $/$  & \gno & 300\\
RCAN \cite{RCAN}   & \gok & \gok & $/$ & \gno & \gno & $/$  & $/$  & $/$  & \gno & 1.5k\\
Ranksrgan \cite{Ranksrgan}  & \gok & \gok & \gno & \gno & \gok & $/$  & $/$  & $/$  & \gok & 276\\ 
SAN \cite{SAN}   & \gok & \gok & \gno & \gno & \gno & $/$  & $/$  & $/$  & \gno & 577\\
HAN \cite{HAN}     & \gok & \gok & \gno & \gno & \gno & $/$  & $/$  & $/$  & \gno & 179\\
PAN \cite{PAN}     & \gok & \gok & \gno & \gno & \gno & $/$  & $/$  & $/$  & \gno & 325\\
SwinIR \cite{SwinIR}  & \gok & \gok & \gno & \gno & \gno & $/$  & $/$  & $/$  & \gno & 5.3k\\
Real-ESRGAN \cite{Real-ESRGAN}  & \gno & \gno & \gno & \gno & \gok & $/$  & $/$  & $/$  & \gno & \textbf{34.1k}\\
IPT \cite{IPT}  & \gok & \gok & \gno & \gno & \gno & $/$  & $/$  & $/$  & \gno & $/$\\
Restormer \cite{Restormer}  & \gok & \gok & \gok & \gno & \gno & \gno & $/$  & $/$  & \gno & 2.4k\\
Uformer \cite{Uformer}  & \gok & \gok & \gno & \gno & \gno & \gno & $/$  & $/$  & \gno & 926\\
HAT \cite{HAT}    & \gok & \gok & \gok & \gno & \gno & \gno & \gno & $/$ & \gno & 1.5k\\
Diffir \cite{Diffir}    & \gok & \gok & \gok & \gok & \gno & \gok & \gok & \gok & \gno & 595\\
Promptir \cite{Promptir}    & \gok & \gok & \gno & \gno & \gno & \gno & \gno & \gno & \gno & 490\\
DiffBIR \cite{DiffBIR}  & \gok & \gok & \gok & \gok & \gno & \gok & \gok & \gok & \gno & 4k\\
X-restormer \cite{X-restormer}  & \gok & \gok & \gno & \gno & \gno & \gno & \gno & \gno & \gno & 135\\
SeeSR \cite{SeeSR}     & \gok & \gok & \gok & \gok & \gok & \gok & \gok & \gok & \gok & 610\\
SUPIR \cite{SUPIR}    & \gok & \gok & \gok & \gno & \gno & \gok & \gok & \gok & \gok & 5.4k\\
HYPIR \cite{HYPIR}  & \gok & \gok & \gok & \gno & \gok & \gok & \gok & \gok & \gok & 1.1k\\
FoundIR-v2 \cite{FoundIR-v2}  & \gok & \gok & \gok & \gno & \gno & \gok & \gok & \gok & \gno & 21\\
\midrule
\textbf{Adoption Rate*} & 96.2\% & 84.6\% & 50.0\% & 14.3\% & 15.4\% & 54.5\% & 66.7\% & 75.0\% & 19.2\% \\
\bottomrule
\end{tabular}
\end{adjustbox}
    \vspace{-4mm}
\end{table*}

\tablename~\ref{tab:iqa-adoption-final} summarizes the IQA metrics adopted by representative image restoration methods from 2014 to 2025.
It can be observed that PSNR and SSIM remain almost universally reported, even though they are widely acknowledged to be insufficient for capturing perceptual or semantic quality.
In many cases, they function more as a form of legacy reference rather than decisive evidence of perceptual superiority.
With the emergence of newer IQA approaches, such as learning-based no-reference metrics, researchers have shown an increasing willingness to incorporate these tools into their evaluation protocols.

At the same time, these metrics introduce new concerns, particularly their tendency to overfit to high-frequency signals.
As a result, artificially enhanced sharpness, exaggerated textures, or irregular high-frequency patterns may be rewarded with higher scores, even when they do not correspond to perceptual quality improvement.

Another notable trend concerns human evaluation.
Across a large portion of the literature, user studies are absent, leaving quantitative metrics as the primary means of demonstrating performance.
Only in more recent years, particularly after 2024, has the frequency of human-centered evaluation begun to increase in several influential works, suggesting a gradual rebalancing of evaluation practices.

A representative example is Real-ESRGAN \cite{Real-ESRGAN}, which did not center its claims around IQA scores; even NIQE was reported only in the appendix. 
Despite this, the model proved highly impactful and widely adopted, accumulating over 34.1k stars on GitHub, demonstrating that long-term practical success and perceived visual quality do not necessarily correlate with benchmark metric dominance.

Overall, these observations suggest that while IQA remains an important tool, the core value of restoration research ultimately lies in improving human-perceived visual quality rather than optimizing for leaderboard metrics.


\section{Human-Centric Evaluation as the Standard}

\label{sec:human}
\subsection{Evolution of Human Evaluation Protocols}

Human assessment has progressed from simple rating scales \cite{sheikh2006statistical} and mean opinion scores (MOS) to more sophisticated methods such as pairwise comparisons, Swiss-style tournaments \cite{ponomarenko2015image}, and ranking systems like ELO scoring \cite{pipal}.  
The key challenge is not whether to incorporate human evaluation, but how to do so in a systematic, reproducible, and scalable manner.
Standardized protocols for human-centered evaluation should be a central focus, enabling fair method comparisons and integration of perceptual feedback into future IQA design.
By combining the strengths of both human judgment and quantitative evaluation, future IQA research can ensure that image quality assessment remains both scientifically rigorous and perceptually meaningful.


\subsection{The Need for Multi-dimensional Human Evaluation}

Human perception of image quality is inherently multifaceted, and reducing evaluation to a single scalar score discards critical information. 
When diverse images are combined into one test set and evaluated jointly~\cite{div2k,isrgenqa}, conclusions often collapse to coarse claims such as “method A outperforms method B.” 
Yet different semantic scenarios pose distinct restoration challenges and emphasize different perceptual factors. 
Performance on faces, text, natural scenes, or structured urban content may reflect fundamentally different model behaviors. 
A method strong in texture-rich scenes may still struggle with facial fidelity or geometric consistency.%

\figurename~\ref{fig:supir_vs_hat} presents a scenario-wise user study comparing SUPIR and HAT. %
Considering only the overall preference, SUPIR appears superior, with $\text{SUPIR} \ge \text{HAT} = 83.5\%$. 
However, scenario-aware evaluation reveals nuances hidden by aggregated scores. 
In hand-drawn scenes, HAT is preferred, with $\text{HAT} \ge \text{SUPIR} = 77.7\%$, and in cartoon/comic scenes, $\text{HAT} \ge \text{SUPIR} = 66.6\%$. 
These results indicate that SUPIR, as a diffusion-based model, may distort stylized content or structures. 
Scenario-aware, fine-grained evaluation thus provides a more systematic understanding of model strengths, weaknesses, and tendencies.

As diffusion models increasingly empower image restoration \cite{SUPIR,HYPIR}, quality is no longer just noise removal, smoothness, or high-frequency detail.
Semantic correctness, such as facial structure or architectural plausibility, has become equally critical. 
A single global IQA score cannot disentangle these factors or reveal where a model succeeds or fails.%

\figurename~\ref{fig:multi-dimension} illustrates a new phenomenon emerging in modern generative restoration models, where over-sharpening and over-generation push results away from both visual comfort and the Ground Truth (GT) images.%
Despite this perceptual deviation, metrics such as MUSIQ, MANIQA, and CLIP-IQA often assign these results very high scores, showing a clear mismatch with human judgment.%
Image restoration models have now moved beyond the earlier stage where the primary limitation was the inability to generate sufficient detail.%
Instead, the challenge has shifted toward excessive detail synthesis, including over-sharpening and hallucinated structures.%
Such issues were largely absent before the diffusion era and represent a fundamentally new challenge for the field.%
Current IQA methods have not yet systematically addressed this problem, as they often conflate increased high-frequency variation with improved perceptual quality.%
This further highlights the necessity of more fine-grained and multi-dimensional image quality assessment pipelines.%

\begin{table*}[tbp]
  \centering
  \renewcommand{\arraystretch}{0.7} 
  \scriptsize    
  \vspace{-3mm}
  \caption{Comparison of Recent IQA and Restoration Models.}
  \label{tab:iqa_ir}

  \begin{tabular}{@{} p{0.09\textwidth} p{0.43\textwidth} p{0.22\textwidth} p{0.17\textwidth} @{}}
    \toprule
    \textbf{Model} & \textbf{Datasets (amount of images)} & \textbf{Training Resolution} & \textbf{Architecture} \\
    \midrule
    \textbf{MUSIQ} & PaQ-2-PiQ (120k), SPAQ (11k), KonIQ (10k), AVA (250k) & Multi-scale (Orig, 224, 384) & Transformer (27M) \\
    \addlinespace[0.3em]  
    \textbf{MANIQA} & PIPAL (23k), LIVE (1.2k), CSIQ (0.9k), TID2013 (3k), KADID (10k) & $224 \times 224$ (Crop) & ViT-B/8 + TAB (135M) \\ 
    \addlinespace[0.3em]
    \textbf{CLIP-IQA} & CLIP-400M (Pre-train), KonIQ-10k (8k fine-tune) & Native (ARP) & CLIP-RN50 (~38M) \\
    \addlinespace[0.3em]
    \textbf{Q-ALIGN} & KonIQ, SPAQ, KADID, AVA (236k), LSVQ (28k video) & Fixed $448 \times 448$ & LLaMA-2-7B (7B) \\
    \addlinespace[0.3em]
    \textbf{DeQA} & KonIQ, SPAQ, KADID, PIPAL & Fixed $448 \times 448$ & LLaMA-2-7B (7B) \\
    \midrule
    \textbf{SUPIR} & 20M HQ images, 70k FFHQ-raw, 100k Negative samples & $512 \times 512$ & SDXL + Adapter (3B+) \\
    \addlinespace[0.3em]
    \textbf{HYPIR} & 20M HQ images, 70k FFHQ-raw & $1024 \times 1024$ & Flux-based (12B) \\
    \bottomrule
  \end{tabular}
  \vspace{-5mm}
\end{table*}

Future evaluations should disentangle semantic validity, detail realism, sharpness, naturalness, visual comfort, and stylistic consistency. 
This is both a core requirement of human-centered assessment and essential for evaluating modern generative restoration models. 
Only through such multi-dimensional, human-centered evaluation can we accurately characterize model behavior and avoid oversimplified conclusions from single-score leaderboards. 
Ultimately, human-centered evaluation should construct a structured representation of perceptual judgments, anchoring restoration models to meaningful improvement and guiding the field toward sustainable progress.%

\section{Metrics Still Matter—But Must Evolve}
\label{sec:future_metric}

Acknowledging the limitations of current metrics does not mean dismissing the value of quantitative evaluation.
The concern is that metrics can detach from human perception, reducing evaluation to a numbers game that may mislead the field. 
Current IQA models are inherently limited and cannot capture the full nuance of human judgment. 
Thus, IQA development remains essential and should advance at least as fast as image restoration technologies to ensure quantitative evaluation stays perceptually relevant.

\subsection{The Scale Gap between IQA and Restoration}

A fundamental but under-discussed issue lies in the scale gap between IQA models and modern restoration systems. 
As summarized in \tablename~\ref{tab:iqa_ir}, mainstream IQA approaches are still built upon limited datasets and model scales.%
Classic IQA datasets such as PaQ-2-PiQ \cite{PaQ-2-PiQ}, SPAQ \cite{SPAQ}, KonIQ \cite{KonIQ}, AVA \cite{AVA}, PIPAL \cite{pipal}, KADID \cite{KADID}, CSIQ \cite{CSIQ}, TID2013 \cite{TID2013}, and LIVE \cite{LIVE} were constructed years ago, primarily focusing on synthetic distortions, limited resolution, and relatively simple content distributions.
The scale of these datasets typically ranges from only ten thousand to a few hundred thousand images.
Correspondingly, the architectures of widely used IQA models (e.g., MUSIQ, MANIQA, CLIP-IQA) remain in the tens to hundreds of millions of learnable parameters.%

In contrast, modern restoration models have undergone a paradigm shift. 
Recent generative restoration systems such as SUPIR and HYPIR are trained on tens of millions of high-quality images (e.g., 20M-scale data) and operate at significantly higher resolutions (up to $1024 \times 1024$), with model capacities reaching billions or even over ten billion parameters.
These models do not merely remove distortions but actively synthesize new high-frequency details and semantic structures.
This discrepancy in data scale, visual complexity, and model capacity may be one of the underlying reasons why existing IQA models struggle to reliably evaluate generative restoration results.

Encouragingly, recent efforts such as Q-ALIGN \cite{Q-Align} and DeQA \cite{deqa} begin to narrow this gap by introducing large language model backbones and multimodal priors into IQA.
These approaches exhibit stronger semantic-level image understanding and show the potential to produce more interpretable and behavior-aware assessments.
This trend suggests a promising direction in which IQA models evolve from distortion detectors toward holistic perceptual and semantic evaluators.%

\begin{figure}[tbp]
    \centering
    \includegraphics[width=\linewidth]{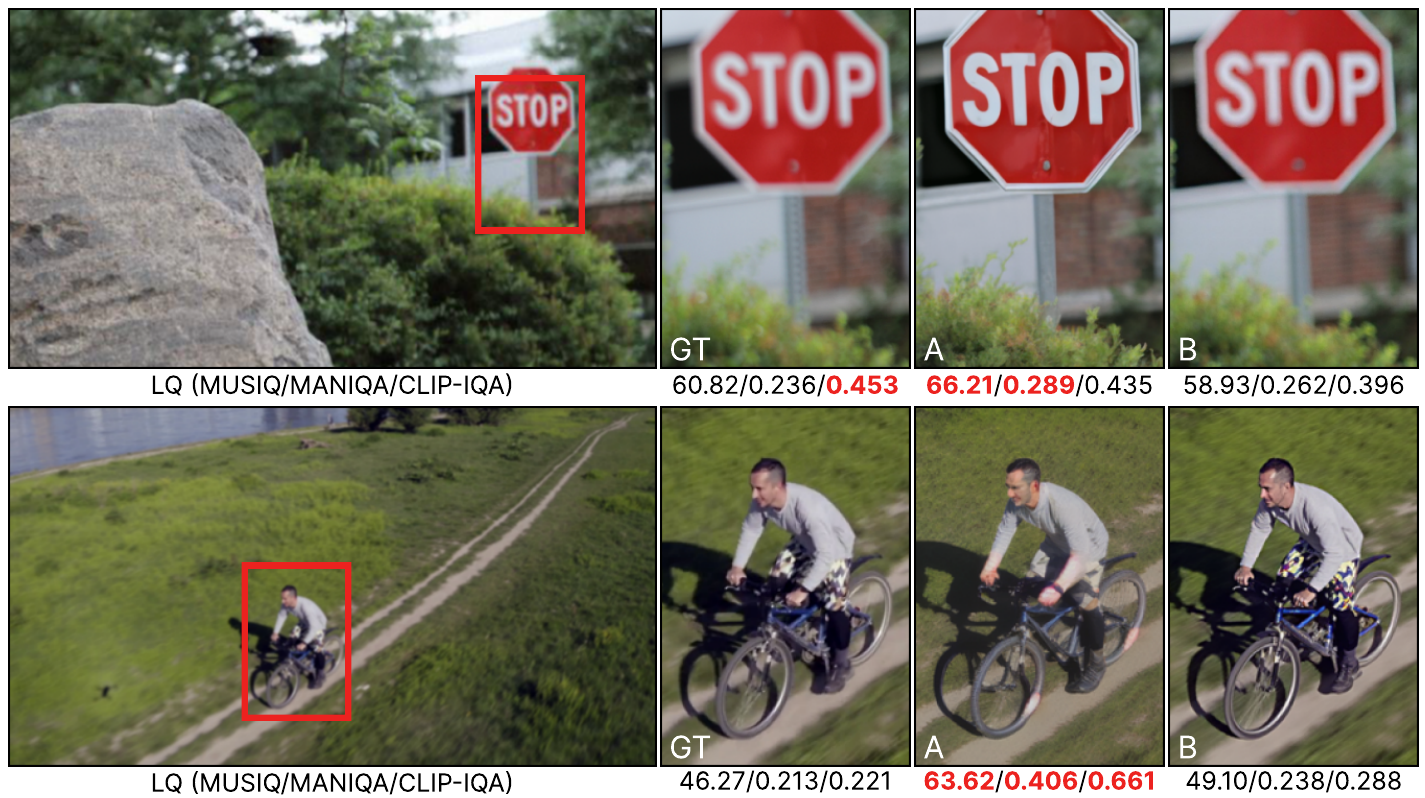}
    \caption{Current IQA metrics tend to favor globally increased high-frequency content, failing to recognize context-dependent blur such as depth-of-field and motion blur. In some cases, no-reference metrics even assign higher scores than GT images.}
    \vspace{-5mm}
    \label{fig:semantic_iqa}
\end{figure}

\subsection{Toward Semantic-Aware Image Quality Assessment}

Bridging the scale gap alone is insufficient if IQA models lack semantic and contextual understanding. 
Future IQA systems should reason about what image content represents and whether visual properties are contextually appropriate, enabling them to distinguish undesirable degradation from intentionally meaningful characteristics. 
For example, depth-of-field naturally blurs backgrounds and motion blur arises from movement; treating such regions as degradations can lead restoration algorithms to over-sharpen them. 
Although this may increase high-frequency content and inflate metric scores, the results often appear perceptually unnatural. 
These issues underscore the need for semantic-aware evaluation that aligns judgments with human perception. %

\figurename~\ref{fig:semantic_iqa} illustrates such failure cases, where current IQA cannot interpret scene content and imaging intent. 
This bias is incompatible with modern image restoration, which involves complex scene structures and semantic objects. 
Without semantic awareness, IQA models cannot judge whether modifications preserve realism or disrupt visual logic. 
Reliable evaluation therefore requires integrating low-level perceptual cues with high-level semantic understanding.

\section{Call to Action}

Reforming our metric-driven culture is a monumental challenge that exceeds the scope of a single paper. Instead of prescribing a rigid framework, we offer the following actionable starting points to catalyze community-wide efforts toward a human-centered evaluation ecosystem:

\textbf{For Reviewers:} Abandon the ``metric-is-king'' culture. Marginal drops in fidelity metrics (e.g., PSNR) should not warrant rejection in generative tasks if a method demonstrates clear advancements in perceptual realism and semantic consistency.
\textbf{For Benchmark Organizers:} Deprecate oversimplified, single-score leaderboards. Benchmarks must adopt scenario-aware protocols, reporting scores segregated by semantic domains and degradation types, to prevent models from hiding domain-specific failures.
\textbf{For Researchers:} Resist artificial tricks (e.g., over-sharpening) aimed solely at inflating IQA scores. Prioritize transparently reporting perceptual trade-offs and elevate human-aligned or VLM-based assessments as primary evidence of performance.
\textbf{For the Community:} Inspired by the Arena \cite{chiang2024chatbot}, we advocate a crowdsourced platform using blind pairwise comparisons and Elo ratings to scale human preference evaluation. The resulting massive data will concurrently serve as invaluable resources for fine-tuning future VLMs as automated proxy evaluators.

It is time for the visual processing community to step out of the comfort zone of outdated metrics and embrace a dynamic, collaborative, and human-centric future.

\vspace{-5mm}
\section{Conclusion}
\label{sec:conclusion}
\vspace{-3mm}
This paper raises a central concern: while evaluation metrics are essential for reproducibility and comparison, they should remain tools grounded in human perception, not the ultimate goal of visual processing research, lest progress become score-driven rather than perceptually meaningful.
Such a shift may steer the field away from its original purpose and favor solutions that look better numerically than perceptually. 
For perception-oriented restoration, assessment should be judged human-centrically, with metrics, user studies, and multi-dimensional analysis guiding models toward generating natural and meaningful results.%

\paragraph{Acknowledgment.} This work was partially supported by the National Natural Science Foundation of China (Grant No.62276251) and  the Joint Lab of CAS-HK. This research was partially funded by the Ministry of Education and Science of Bulgaria (support for INSAIT, part of the Bulgarian National Roadmap for Research Infrastructure).



\bibliographystyle{unsrt}
\bibliography{refs}

\end{document}